# Mining Domain Knowledge: Improved Framework towards Automatically Standardizing Anatomical Structure Nomenclature in Radiotherapy

**Qiming Yang[1,2], Hongyang Chao[1,2], Dan Nguyen[1], and Steve Jiang[1]**

[1]Medical Artificial Intelligence and Automation Laboratory, Department of Radiation Oncology, University of Texas Southwestern Medical Center, Dallas, TX, USA

[2]School of Data and Computer Science, Sun Yat-sen University, Guangzhou, GD, China

Corresponding author: Steve Jiang (e-mail: Steve.Jiang@UTSouthwestern.edu).

**ABSTRACT** The automatic standardization of nomenclature for anatomical structures in radiotherapy (RT) clinical data is a critical prerequisite for data curation and data-driven research in the era of big data and artificial intelligence, but it is currently an unmet need. Existing methods either cannot handle cross-institutional datasets or suffer from heavy imbalance and poor-quality delineation in clinical RT datasets. To solve these problems, we propose an automated structure nomenclature standardization framework, 3D Non-local Network with Voting (3DNNV). This framework consists of an improved data processing strategy, namely, adaptive sampling and adaptive cropping (ASAC) with voting, and an optimized feature extraction module. The framework simulates clinicians' domain knowledge and recognition mechanisms to identify small-volume organs at risk (OARs) with heavily imbalanced data better than other methods. We used partial data from an open-source head-and-neck cancer dataset to train the model, then tested the model on three cross-institutional datasets to demonstrate its generalizability. 3DNNV outperformed the baseline model, achieving higher average true positive rates (TPR) over all categories on the three test datasets (+8.27%, +2.39%, and +5.53%, respectively). More importantly, the 3DNNV outperformed the baseline on the test dataset, 28.63% to 91.17%, in terms of F1 score for a small-volume OAR with only 9 training samples. The results show that 3DNNV can be applied to identify OARs, even error-prone ones. Furthermore, we discussed the limitations and applicability of the framework in practical scenarios. The framework we developed can assist in standardizing structure nomenclature to facilitate data-driven clinical research in cancer radiotherapy.

## I. INTRODUCTION

In the field of radiotherapy (RT), nomenclature standardization is the process of imposing a unified and structured labeling system on anatomical structures [1, 2, 3]. This is a prerequisite for clinical data curation and data-driven research, especially in the era of big data and artificial intelligence [1, 4, 5, 6, 7]. However, because of differences in local policies, vendors, and language environments, structure labels are often inconsistent [8, 9]. A large number of retrospective RT datasets [10, 11] cannot be shared and reused without consistent labels, and manually cleaning RT data is very expensive and time-consuming [8, 9, 12, 13, 14]. Therefore, it is necessary to develop software tools to automate nomenclature standardization to facilitate data-driven clinical research.

Previous works have proposed standardizing the nomenclature of anatomical structures via text-based methods that rely on label matching and clinicians' intervention to correct mismatched labels at a single institution [8, 15, 16]. However, language constantly changes, and different naming conventions make the semantic information in labels difficult to recognize automatically. As a result, text-based methods cannot be applied to datasets collected even from a single institution, let alone to cross-institutional datasets.

The labels of organs at risk (OARs) have a one-to-one correspondence with the images (such as Computed Tomography [CT] scans and segmentation masks), and the image data contain invariant semantic information that can standardize nomenclature in multilingual environments. Methods that leverage this image information to tackle cross-institutional RT datasets are called image-based methods. Existing image-based methods try to automatically standardize nomenclature by exploiting semantic invariance in the image [17, 18, 19, 20]. Among these methods, algorithms that leverage atlas-based registration can also be used to determine the category of the structure and then relabel it [18, 19]. However, atlas-based registration is unstable and time-consuming. Other image-based methods convert the task of structure nomenclature standardization to OAR classification based on deep learning (DL) frameworks [17, 20]. Nonetheless, these methods have largely overlooked the problems caused by imbalance and poor delineation in real RT datasets, especially for small-volume OARs with similar positions, shapes, and sizes, such as the pituitary and optic chiasm. RT datasets are imbalanced not only in the number of OARs but also in the size of each OAR. For example, in Fig. 1 (a), the volume of the brain is much larger than that of the pituitary. Models built on such datasets tend to be biased and inaccurate. Poor delineation of OARs increases the inter-class similarity and the intra-class variation. For example, in Fig. 1 (b), the pituitary and optic chiasm are very similar, but the larynx varies greatly across patients. Both imbalance and poor delineation will bias the classifier, which will lead to incorrect predictions for small-volume OARs.

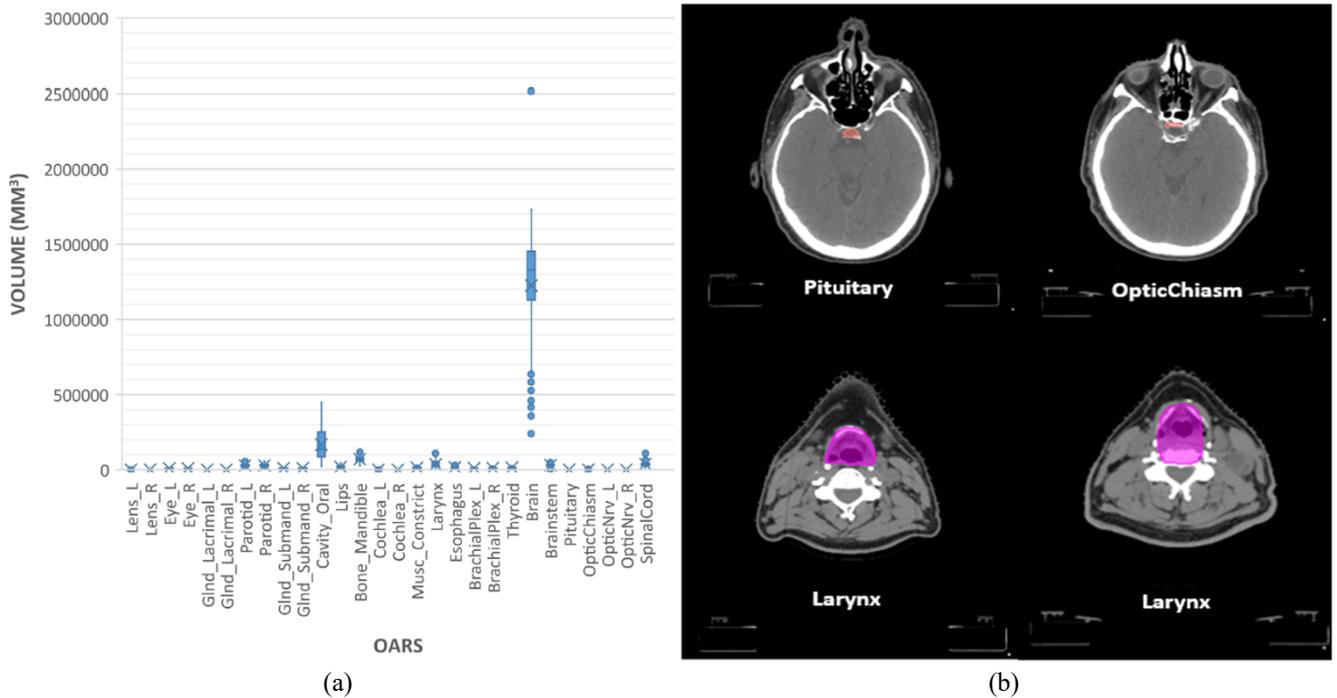

**FIGURE 1.** Characteristics of OARs in real RT data. (a) The size of the OARs is extremely imbalanced. (b) The poor delineation in RT data: the first row indicates the similarity of small-volume OARs (inter-class similarity), and the second row shows examples of poor delineation for the same OAR (intra-class variation).

As mentioned earlier, some image-based methods treat the task of structure nomenclature standardization as an OAR classification task [17, 20]. In the field of computer vision, deep learning has led to a series of breakthroughs for classification tasks [21, 22] that have improved upon traditional classification methods [23, 24]. Related works seeking to improve the performance of DL-based classifiers have mainly focused on three aspects: 1) constructing deeper, wider, and more elaborate architectures [22, 25, 26, 27, 28, 29, 30] to increase the capacity for adapting data and training [31]; 2) enriching samples to get close to the actual distribution [32, 33, 34, 35, 36]; and 3) adding subjective constraints to make high-level features extracted within the network correspond to the domain knowledge required for specific tasks [37, 38, 39, 40]. Existing state-of-the-art networks [21, 22, 25, 26, 27, 28, 29, 30] for classification can be applied to the current task. ResNet [25] has a lower computational cost but better performance than other networks [41]. Therefore, we have made many attempts to use ResNet50 for this application, but these attempts have yielded results similar to previous reports [17, 20]: the true positive rate (TPR) of small-volume OARs (such as the pituitary and optic chiasm) cannot meet the requirements for clinical implementation. It is worth noting that clinicians can make quick and accurate decisions for small-volume OARs, even with poorly delineated samples, which means the images contain enough effective information for clinicians to apply their domain knowledge and recognition mechanisms. To date, there has been no relevant research on how clinicians make accurate decisions when classifying OARs, but we can simulate this process and, thus, incorporate the implicit domain knowledge and recognition mechanisms necessary for decision making into the target framework.

The main goal of our work is to explore ways to integrate clinicians' domain knowledge and recognition mechanisms into a neural network to improve the classifier's performance for categorizing small-volume OARs. To this end, we propose an automatic structure nomenclature standardization framework, 3D Non-local Network with Voting (3DNNV). This framework consists of an improved data processing strategy and an optimized feature extraction structure. The data processing strategy was proposed to provide the explicit information that clinicians use when labeling structures. The feature extraction structure simulates the observation process, which enhances the observational fineness of the region of interest (ROI) in the high-level features.

*A. Improved data processing strategy*

We propose a simple and effective adaptive data processing strategy: adaptive sampling and adaptive cropping (ASAC) with voting. ASAC simulates the process of clinicians observing images and collecting the information needed for decision making, and it generates multi-scale and multi-position inputs for a sample. ASAC constructs a set of augmented inputs, assists the model in mining the effective information implicit in the raw data, and extracts the domain knowledge that clinicians typically need to identify OARs. The voting strategy accounts for variations in a structure's shape and location that may lead to poor delineation. This strategy is a weighted sum of all the predictive results of inputs for the same sample; this makes the final result closer to predicting the "ideal" semantic features. The voting strategy also agrees with the principle of clinicians making decisions based on comprehensive information.

*B. Optimized feature extraction structure*

The convolutional network only processes one local neighborhood at a time, and the common way to model the long-range dependency on semantic features is to increase the receptive field. In order to fill the gaps in capturing long-range dependency and to enhance the observational fineness in the region of interest in the high-level semantic features, we added non-local blocks [42] to ResNet50 to optimize the feature extraction structure in the network (designated "NN" for Non-local Network). Non-local blocks apply a self-attention mechanism [43] to image sequence processing by calculating the similarity matrix for high-level semantic features, thereby containing the long-range dependency and enhancing the representation of the semantic features.

By combining the ASAC/Voting strategy with the Non-local Network, we obtained the final framework, 3DNNV, which can standardize the nomenclature of structures in RT datasets. The 3DNNV integrates clinicians' domain knowledge and recognition mechanisms into the final model from a new perspective, mitigates the problems caused by imbalance and poor delineation in RT datasets, and improves the performance for identifying small-volume OARs. This framework allows us to categorize structures in cross-institutional RT data quickly and efficiently, then automatically relabel these structures with general labels recommended in AAPM TG-263 [1]. Furthermore, 3DNNV is extensible and can be easily transferred to other anatomic sites after fine-tuning on a few samples.

The rest of this paper is organized as follows: Section II introduces related works that have sought to automate nomenclature standardization of OARs in recent years. Section III describes our 3DNNV framework. Section IV shows the results of experiments evaluating 3DNNV's performance and comparing it with other state-of-the-art methods. Section V discusses the limitations of this study and the future prospects of our work. Section VI summarizes our main findings and provides future directions.

TABLE I
EXAMPLES OF INCONSISTENT LABELS IN RT DATASETS. THERE ARE VERY DIFFERENT LABELS FOR THE SAME OARs, SUCH AS "NOD" AND "NERF OPT DRT."

| Patient 1 | Patient 2 | Patient 3 |
|---|---|---|
| Parotide D | RT Parotid | PAROTIDE D |
| GTVggIIID | Mandible | MANDIBULES |
| Ext 0.5 | LT Eye | PLEXUS BR D |
| nerf opt drt | Cord | NOD |
| Moelle | LT SUBMANDIBULAR | GL LACR D |
| Oeil gche | GTV | CERVEAU-NUQUE |
| Cerveau | CTV1 | CONF_I PTV3 5412(1DMPO1.1)_1 |
| External | EXTERNAL | |
| pt pr | Midline | PEAU |
| Tronc cerebral | Brainstem | CRISTAL G |
| Cristallin G | LT Parotid | |

## II. RELATED WORK

*A. Text-based methods*

Text-based methods standardize structure nomenclature mainly by using structured naming templates or label mapping dictionaries. Mayo et al. [15] built software containing structured templates, which allows clinicians to relabel structures interactively. The fixed template helps to unify labels better than free-text interactive tools. Nyholm et al. [16] mapped the main structure labels in local clinical centers to the name list of the general naming convention, then manually corrected the mismatched labels through the interactive interface. The authors used the tool to aggregate RT data from 15 medical centers in Sweden. More recently, Schuler et al. [8] pointed out that, when standardizing radiotherapy data, it is difficult to distinguish between typographic name variations and fundamental semantic differences in the same structure.

Therefore, they developed a tool called Stature that maps a local standard structure name (LSSN) to the AAPM TG 263 naming table by creating a lookup dictionary. The above methods map the original labels to standardized labels based on a dictionary and manual intervention. These kinds of methods can establish the mapping between the original labels and standardized labels to quickly solve the problem of inconsistent labels in the local RT dataset. However, language constantly changes, as shown in Table I, which limits these methods' applicability to cross-institutional datasets. In addition, the text-based methods cannot handle large-scale retrospective datasets.

*B. Image-based methods*

Image-based methods, which are based on the invariant semantic information in medical images, are learnable automatic recognition methods that overcome the problems inherent in text-based methods. The label propagation, which is implemented by an atlas-based deformable image registration (DIR) algorithm, registers an atlas with known labels to the input and then chooses the one with the highest overlap mask to relabel the input [19]. In this way, unknown datasets can be standardized by labels in the atlas. However, the DIR's performance is unstable [18]. Also, this method is highly time-consuming, so it falls well short of practical requirements.

Our previous work departed from these methods, as it converted label standardization to the task of automatically categorizing structures in RT data and modeled the process with a deep neural network, which used the weighted mask of OARs to construct a composite mask as 2D input [17]. This work demonstrated the excellent performance of deep learning networks in standardizing OAR labels, but the experiment did not make full use of the three-dimensional shape and location information on the CT. The classes of OARs in the training dataset were clean and sufficient, but the real dataset contained many other challenges, such as heavy data imbalance, inter-class similarity, and intra-class variation, that could limit the method when extending it to other anatomic sites. More recently, Rhee et al. [20] extended the number of categories to 19 OARs in the head-and-neck region and loosely utilized the encoder of V-Net [44] to construct their framework, TG263-Net. This framework leveraged 3D inputs and achieved high accuracy in identifying 19 OARs, but it did not take into account imbalance and poor delineation in RT datasets, so its performance in identifying small-volume OARs is insufficient for practical clinical needs.

## III. MATERIALS AND METHODOLOGY

*A. Overview of 3DNNV*

This section outlines the workflow of 3DNNV (Fig. 2). 3DNNV consists of two parts in the inference phase for standardizing structure nomenclature: 1) the ASAC/Voting strategy and 2) the Non-local Network. For any OAR in given Digital Imaging and Communications in Medicine (DICOM) data, the CT and corresponding mask are extracted to form a raw data pair. Then, ASAC generates multi-scale and multi-position inputs for each sample. During training, each input generated by ASAC is regarded as an independent sample, and the parameters of the non-local network are updated and optimized based on the samples in each mini-batch. In the inference phase, multiple inputs for a sample are fed into the network, which outputs the vectors (Vectors in Fig. 2, 256-d vector for each). Sharing weights here allows the consistent representation of multiscale/multi-position inputs in feature space so that we can leverage the same model with the same parameters to extract high-level features for each input. The 256-d vectors vote for a final predictive result as the output of 3DNNV, and the sample is renamed with a standardized label.

*B. Data*

In accordance with Brouwer et al.'s suggestion [45], we selected the 28 categories of head-and-neck OARs shown in Fig. 1 (a) to train our model. We compared our model's performance in standardizing structure nomenclature against other models by testing them on three different head-and-neck image datasets.

1) HN_PETCT

HN_PETCT [46, 47] is an open-source head-and-neck RT dataset released on The Cancer Imaging Archive (TCIA) [48] that includes data collected from four different French medical institutions comprising 298 patients. We collected 4372 samples in total for the 28 OAR categories. Then, we divided the samples into three subsets for training, validation, and testing in a ratio of 3:1:1. It should be noted that the number of samples in the dataset is extremely imbalanced. For Glnd_Lacrimal_L/R and Pituitary, only 9 samples for each were used as training data.

2) PDDCA

PDDCA [49] is an open-source RT dataset containing data from 48 patients that was released by the MICCAI 2015 Segmentation Challenge. This dataset contains only 9 categories of head-and-neck OARs (Parotid_L, Parotid_R, Glnd_Submand_L, Glnd_Submand_R, Bone_Mandible, Brainstem, OpticChiasm, OpticNrv_L, and OpticNrv_R). All contours for OARs were re-delineated by trained radiologists. We collected 408 samples in total and used all of them as a test set.

3) HN_UTSW

HN_UTSW is an RT dataset collected by our team at UT Southwestern that contains data for 408 patients. We collected a total of 5153 samples for 28 OARs (the same categories as HN_PETCT) and used all of them for testing to show our model's generalizability.

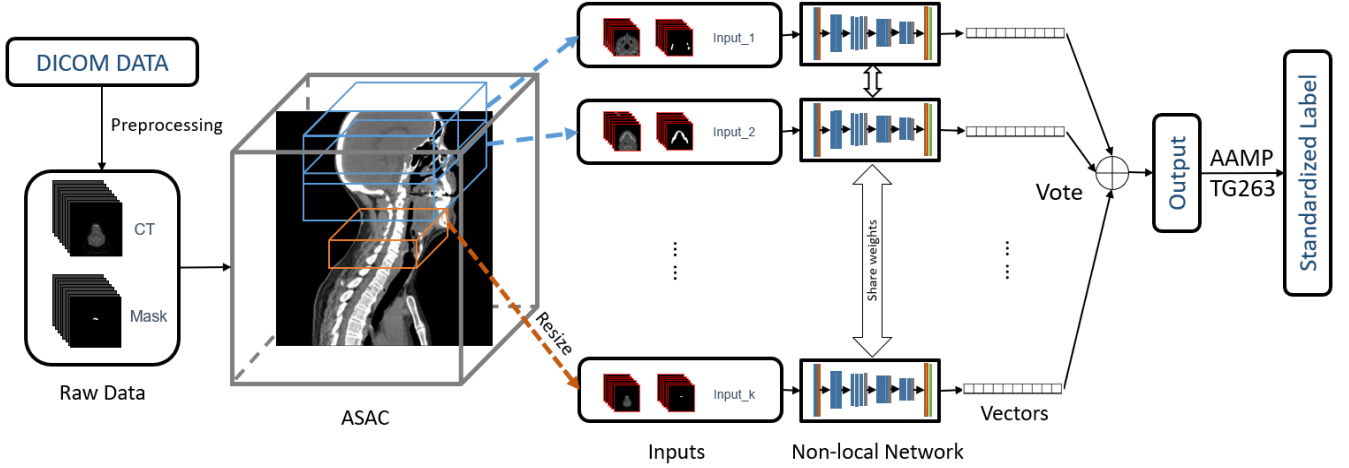

**FIGURE 2.** Overview of 3DNNV in the inference phase.

## C. Preprocessing

For each patient's data in given DICOM files, 3D CT volumes and corresponding masks were extracted to form raw data, then the voxel size of the 3D volumes in the raw data was normalized. To ensure that the small-volume OARs do not lose any information through down-sampling, we chose to use the same voxel dimension ratio z:y:x = 0.77:1:1 from the training dataset HN_PETCT for all other datasets. We performed trilinear interpolation for resizing and reshaping. Due to differences in maxima (and minima) of Hounsfield unit (HU) values for different patients, the range of HU values was truncated to [-1000, 2500], then normalized to [0, 1]. We directly used a binary [0, 1] matrix to represent the mask. For some patients, the OAR contours may be missing in some intermediate slices and then are generated using the nearest slices.

## D. 3DNNV: 3D Non-local Network with Voting

ASAC/Voting is an essential part of 3DNNV. It is worth noting that ASAC is a data processing strategy that can be applied in all stages, but the voting strategy is applied only in the inference phase.

### 1) ASAC: ADAPTIVE SAMPLING AND ADAPTIVE CROPPING

For each OAR, a pair of pre-processed 3D CT and mask volumes (Raw Data in Fig. 2) are cropped into smaller volumes using sliding cubes of $n \times m \times m$ (the blue and orange cubes in the part of ASAC in Fig.2) along the patient long axis, which are then used as inputs for the non-local network. Here $m$ and $n$ are the sizes of the sliding cube in axial plane and in patient long axis direction. In our experiments, we use 5 different sizes for the cropping cubes: 12×128×128, 18×192×192, 24×256×256, 30×320×320, and 36×384×384. The cubes slide at a step size of $n/3 \times 2$. The cropped image volumes using cubes of different sizes are resized into 12×128×128 before being inputted into the non-local network. ASAC is not only a way to extract clinicians' domain knowledge but also a way to deal with the issues related to limited computational resources and imbalanced training datasets. For some oversized OARs, such as Brain, the contour in the axial slice cannot be entirely captured by small-volume cubes (such as 12×128×128). Therefore, it is necessary to adaptively resize the CT and mask first to fit them into the cubes, then perform the sampling. By performing ASAC, we gained multi-scale and multi-position inputs for each sample.

### 2) VOTING

As mentioned above, ASAC generates multi-scale and multi-position inputs, which contain global and local information. The outputs of the non-local network, corresponding to different inputs, will be summed up to vote for the final recognition result. This voting strategy is used at the inference phase.

### 3) NON-LOCAL NETWORK

We set vanilla ResNet50 as the backbone for our 3DNNV network and also as the baseline for our performance comparisons (Table II).

Then, we added non-local blocks [42] to the backbone network to form the final 3D non-local network (Fig. 3). Inspired by the self-attention mechanism [43], Wang et al. [42] proposed the non-local block to capture the global dependence on semantic features. It was designed to handle sequential data, so we stacked it into our framework. In this work, we are committed to enhancing the position information's dependence on the CT image and the shape information's dependence on the mask image, so the pairwise function $f$ may be implemented using the concatenated form. The non-local block used in our network is defined as follows:

$$y_i = \frac{1}{C(x)} \sum_{\forall j} f(x_i, x_j) \psi(x_j) \quad (1)$$

$$f(x_i, x_j) = ReLU(W_f^T[\theta(x_i), \mu(x_j)]) \quad (2)$$

$$z_i = \sigma(y_i) + x_i \qquad (3)$$

$x$ and $z$ are set as the input and output, respectively, of the non-local block. Both are of the same size: $B \times C \times D \times H \times W$. $B$ denotes the batch size of the input, and $C$ represents the number of channels. $D$, $H$, and $W$ are depth, height, and width, respectively. Here, $i$ is the index of an output position whose response is to be computed, $j$ is the index of all possible positions, and $y$ is an intermediate output with the same size as $x$. $\psi$, $\theta$, $\mu$, and $\sigma$ are all $1 \times 1 \times 1$ convolution layers. Operator [.,.] indicates the concatenation operation, and $W_f$ is the mapping matrix that converts the concatenated vector to the scalar output. "$+x_i$" indicates identity mapping, and the input $x_i$ is added to the transformed $y$ to get the final output $z$ of the non-local block. $C(x)$ is a regularization term: $C(x) = D \times H \times W$.

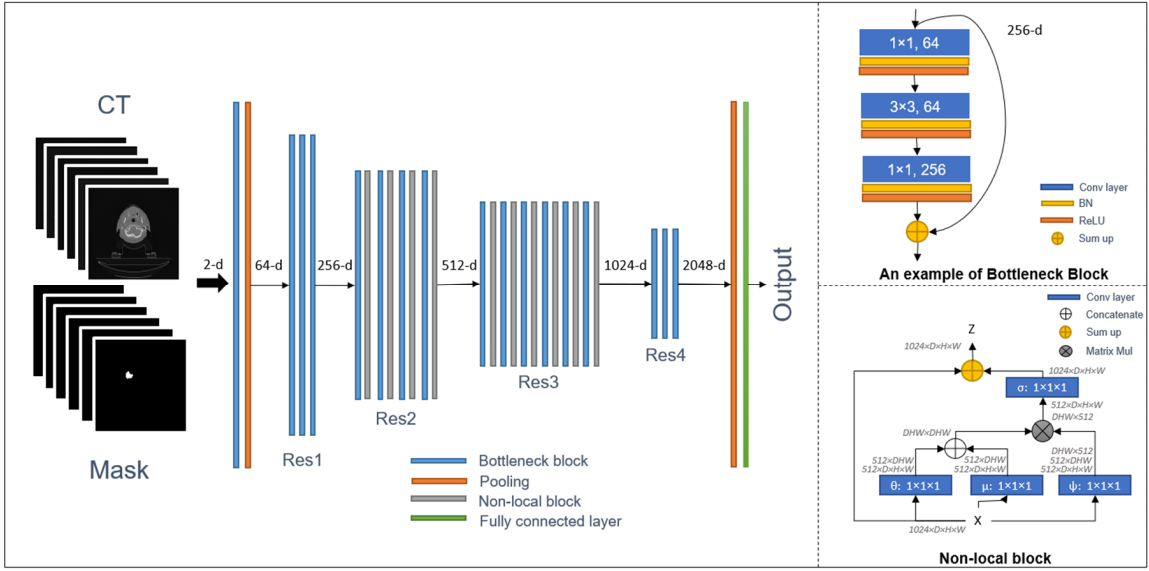

**FIGURE 3.** Non-local Network. To construct the Non-local Network in our work, we stacked one Non-local block at the end of each Bottleneck block in Res2 and Res3. Details for the Bottleneck block and Non-local block are shown in the figure. The self-attention mechanism [14] is applied in the Non-local block to capture the long-term dependency within semantic features. Any input for the network is 2-channel 3D data (12×128×128 CT and 12×128×128 mask), and the corresponding output will be 256-d vector.

## IV. EXPERIMENTS

### A. Experimental setting

#### 1) TRAINING DETAILS

Using the training data outlined in section III.B and the preprocessing outlined in section III.C above, we trained our deep learning models as described below. To account for imbalance in the numbers of images for each OAR in the dataset, we applied a non-uniform sampling method—OARs that were represented less were inversely proportionally sampled more times. We augmented the training data by performing affine transformations, including randomly translating, rotating, shearing, and scaling. Finally, the central cube of the sample was cropped as input data. The final input data size was $2 \times 12 \times 96 \times 96$, which is two-channel 3D data that includes the 3D CT volume and the corresponding mask on the same slices. All architectures used in this work were initialized as described by He et al. [51]. The Adam optimization algorithm [52] was applied to optimize the networks with an initial learning rate of 1e-4, and cross-entropy was set as the loss function. The batch size was set to 16. For samples generated by ASAC, we set the total number of epochs to 20, and the learning rate dropped by a factor of 10 after 2, 5, and 10 epochs. For other architectures without ASAC, we set the total number of epochs to 200, and the learning rate decreased by a factor of 10 after 10, 20, and 30 epochs. The 3DNNV was implemented on the PyTorch1.0 [50] framework and trained on a single GPU NVIDIA Tesla K80.

#### 2) EVALUATION

For this multi-class classification task, we used true positive rate (TPR), F1 score, and area under the receiver operating characteristic curve (AUC) to evaluate the performance of our models. These metrics are defined as follows:

$$\text{TPR} = \frac{TP}{TP+FN} \qquad (4)$$

$$\text{PPV} = \frac{TP}{TP+FP} \qquad (5)$$

$$\text{F1} = 2 \cdot \frac{PPV \cdot TPR}{PPV+TPR} \qquad (6)$$

$$AUC = \frac{\sum_{ins_i \in positive} rank_{ins_i} - \frac{M \times (M+1)}{2}}{M \times N} \quad (7)$$

Multi-class classification can be considered as multiple binary classifications and can calculate true positive (TP), false negative (FN) and false positive (FP) values for each category separately. F1 score is the harmonic mean of the positive predictive value (PPV) and TPR. In (7), $rank_{ins_i}$ means the *i*-th positive sample sorted by probability. AUC indicates how well the model distinguishes between different classes. AUC is not sensitive when used on an imbalanced test sample.

### B. Comparisons among ResNet models

We developed the 3DNNV model in a step-wise manner. First, we set vanilla 3D ResNet50 as the backbone network; then, we optimized the architecture; and finally, we integrated domain knowledge into the network. We evaluated and compared the performance of the models obtained at each step.

Our first goal was to determine an initial preprocessing strategy for the raw data. Beginning with the baseline network, we tested three different strategies: taking global samples without voxel size normalization (GS), global samples with voxel size normalization (VN_GS), and local samples with voxel size normalization (VN_LS) as inputs. Samples collected at the scale of 36×384×384 were marked as global samples (GS), and samples collected at the scale of 12×128×128 were marked as local samples (LS). "VN" means voxel normalization. Accordingly, we designated the architectures as Baseline (GS), Baseline (VN-GS) and Baseline (VN-LS). We found that, for the error-prone small-volume OARs in the head-and-neck region, detailed information contained in the local sample plays an important role in recognition (see Fig. 4 and Table II), so incorporating local details benefits models in classifying small-volume OARs. However, the model trained only on local samples, Baseline (VN-LS), could not distinguish between BrachialPlex_L (BP_L) and BrachialPlex_R (BP_R) (Fig. 5). Without the global location information, the model failed to indicate on which side the OAR should be.

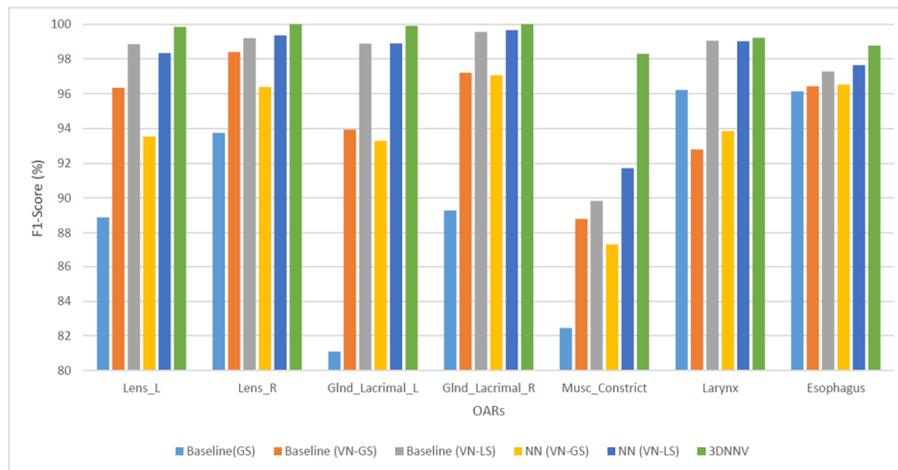

FIGURE 4. Average F1 Scores (%) for ResNet-based models' classification of small-volume OARs in HN_UTSW.

To enhance the representation of small-volume OARs in high-level feature space, we added non-local blocks to the backbone networks with voxel normalization and compared the performance of the non-local network (NN) with the baselines. We designated the non-local network architectures as NN (VN-GS) and NN (VN-LS). The NN architectures performed slightly better than the baselines over all categories, especially for Pituitary and OpticChiasm (Table II). However, their performance on other small-volume OARs was barely satisfactory. Like the Baseline (VN-LS) architecture, the non-local network architecture trained on local samples, NN (VN-LS), could not distinguish between BrachialPlex_L (BP_L) and BrachialPlex_R (BP_R).

Finally, we applied the ASAC/Voting strategy to generate multiple inputs for a sample and combine the information through voting. We constructed and trained the 3DNNV network on the samples generated by ASAC. In the inference phase, all output vectors for the same sample voted for the final predictive result. We found that 3DNNV performed well in identifying small-volume OARs, even those similar in shape, size and location, such as the pituitary and optic chiasm.

When we compared the performance of the six ResNet-based models in classifying the 28 OARs across all three institutional test sets, we found that 3DNNV was superior to the baseline methods for classifying OARs and had good generalizability across different institutional datasets (Table III).

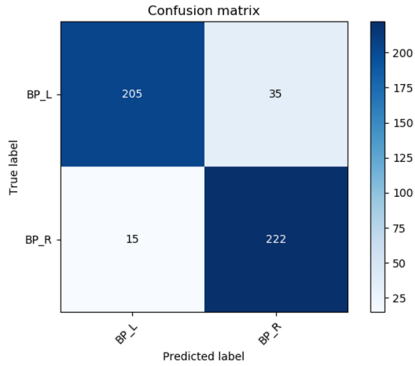

**FIGURE 5.** Confusion matrix of Baseline (VN-LS) on BrachialPlex_L and BrachialPlex_R.

TABLE II
AVERAGE F1 SCORE (%) OF RESNET-BASED MODELS FOR PITUITARY AND OPTICCHIASM ON HN_UTSW.

| Architecture | Pituitary | OpticChiasm |
|---|---|---|
| Baseline (GS) | 28.63±8.33 | 86.35±4.06 |
| Baseline (VN-GS) | 61.87±18.73 | 96.83±1.91 |
| Baseline (VN-LS) | 42.61±10.87 | 90.41±2.38 |
| NN (VN-GS) | 82.05±15.77 | 98.32±0.97 |
| NN (VN-LS) | 59.26 ± 8.83 | 95.41 ± 1.73 |
| 3DNNV | **91.17±3.18** | **99.47±0.18** |

TABLE III
EVALUATION OF BASELINE, NN AND 3DNNV MODELS ON THREE DIFFERENT TEST DATASETS.

| Architecture | HN_PETCT | | | PDDCA | | | HN_UTSW | | |
|---|---|---|---|---|---|---|---|---|---|
| | TPR (%) | F1 (%) | AUC (%) | TPR (%) | F1 (%) | AUC (%) | TPR (%) | F1 (%) | AUC (%) |
| Baseline (GS) | 91.54±17.13 | 91.61±15.96 | 95.72±8.59 | 97.61±5.02 | 98.51±3.22 | 98.79±2.55 | 93.36±9.61 | 91.81±13.28 | 96.60±4.82 |
| Baseline (VN-GS) | 95.33±10.71 | 95.83±9.00 | 97.64±5.36 | 99.04±2.08 | 99.43±1.10 | 99.51±1.03 | 96.54±5.29 | 95.75±7.13 | 98.23±2.65 |
| Baseline (VN-LS) | 98.56±2.49 | 98.72±1.93 | 99.26±1.25 | 97.79±6.40 | 98.69±3.73 | 98.89±3.20 | 96.03±6.04 | 93.96±12.18 | 97.96±3.07 |
| NN (VN-GS) | 96.07±7.97 | 96.27±7.36 | 98.00±3.99 | 98.96±3.12 | 99.45±1.65 | 99.48±1.56 | 96.56±5.15 | 96.53±4.05 | 98.24±2.58 |
| NN (VN-LS) | 98.71±1.71 | 98.64±1.54 | 99.34±0.85 | 99.05±2.22 | 99.51±1.16 | 99.53±1.11 | 96.65±5.20 | 95.57±8.31 | 98.27±2.64 |
| 3DNNV | **99.81±0.63** | **99.82±0.39** | **99.90±0.31** | **100±0** | **100±0** | **100±0** | **98.87±2.37** | **98.90±1.91** | **99.42±1.19** |

*C. Comparisons with previous works*

To further test 3DNNV's ability to standardize structure nomenclature, we compared its performance with that of other image-based methods: specifically, atlas-based registration and several deep learning-based methods.

1) ATLAS-BASED REGISTRATION

Atlas-based registration can standardize structure nomenclature by matching OARs with an atlas in the database and renaming the input with the atlas label that has the largest overlap mask. To test atlas-based registration for this application, first, we constructed a 2D single-atlas database for the 28 OARs, each sample in which contained a CT slice and a mask for the OAR to be identified in the same slice. Second, for each pair of CT and mask of the OAR to be identified (noted as fixed CT and fixed mask), the moving CT in each atlas was registered to the fixed CT, and the transformation (warping parameters) was learned. Applied the transformation on the moving mask of atlas, and then, the area of overlap between the deformed moving mask and the fixed mask was calculated by using the Dice Similarity Coefficient (DSC). DSC is shown in the following formula, with $X$ and $Y$ denoting the given fixed mask in given data and the moving masks in the atlas database.

$$DSC = \frac{2|X \cap Y|}{|X|+|Y|} \quad (8)$$

For the experiment comparing atlas-based registration with 3DNNV, every structure processed by 3DNNV was first run through an early-match module to avoid processing standardized structures repeatedly. The early-match module performed string matching between the original label and the standardized label: if and only if the original label fully matched one of the standardized labels in the dictionary, then the original label was treated as an already standardized label. Otherwise, the structure was fed into 3DNNV to obtain the prediction result. To limit the running time, we tested both methods on data from two randomly selected patients in the HN_UTSW dataset (Table IV).

TABLE IV
TESTING RESULTS OF ATLAS-BASED REGISTRATION AND 3DNNV ON DATA FROM TWO PATIENTS

| Evaluation (Positive/All, Time) | Patient 1 | Patient 2 |
|---|---|---|
| Atlas-based registration | 13/14, 19m 34.25s | 7/24, 32m 35.67s |
| **3DNNV w/ early-match** | **14/14, 32.94s** | **23/24, 1m 19.86s** |

The results show that the atlas-based registration algorithm is very time-consuming and unstable on different patient datasets, and its running time is almost 30 times longer than 3DNNV's, which makes atlas-based registration unacceptable for this application. The registration effect of atlas-based deformable image registration often depends on the atlas dataset, the deformation model, and the objective function. However, it is difficult to construct an optimal single-atlas database. Multi-atlas datasets could be applied to make up for this deficiency, but this would be even more time-consuming.

2) DL-BASED METHODS

We also applied and analyzed other DL-based methods for structure nomenclature standardization and compared their performance with 3DNNV. Taking into account the massive impact of different sampling strategies and networks on performance, we set several different architectures for the experiment. For the various inputs used in this section, 1c2d is a 1-channel composite mask [17], 2c2d is a 2-channel input combining 2D CT and the corresponding mask, and 2c3d is a 2-channel 3D CT and mask [20]. For the different networks, we trained and tested 5-layer CNN [17], vanilla 2D ResNet50 [25], and TG263-Net [20] on the same datasets and compared their performances with 3DNNV. To fairly compare different methods with different inputs, we set 4 architectures—5-layer CNN (1c2d), ResNet50 (1c2d), 5-layer CNN (2c2d) and ResNet50 (2c2d)—to determine the best combination of network and inputs. We found that the ResNet50-based models performed far better than the 5-layer CNN models (Fig. 6, Table V, and Table VI), even though ResNet50 has fewer parameters and costs less on computation. The 2-channel inputs include more information, which generally improves the overall performance of 28 categories (Table VI). Of note, Pituitary got an F1 score of 0.0% (Table V) because of the extremely imbalanced training sample: not only were there many more samples for Optic Chiasm than for Pituitary, but the two OARs are quite similar and error-prone. As a result, all test samples for Pituitary were predicted as Optic Chiasm.

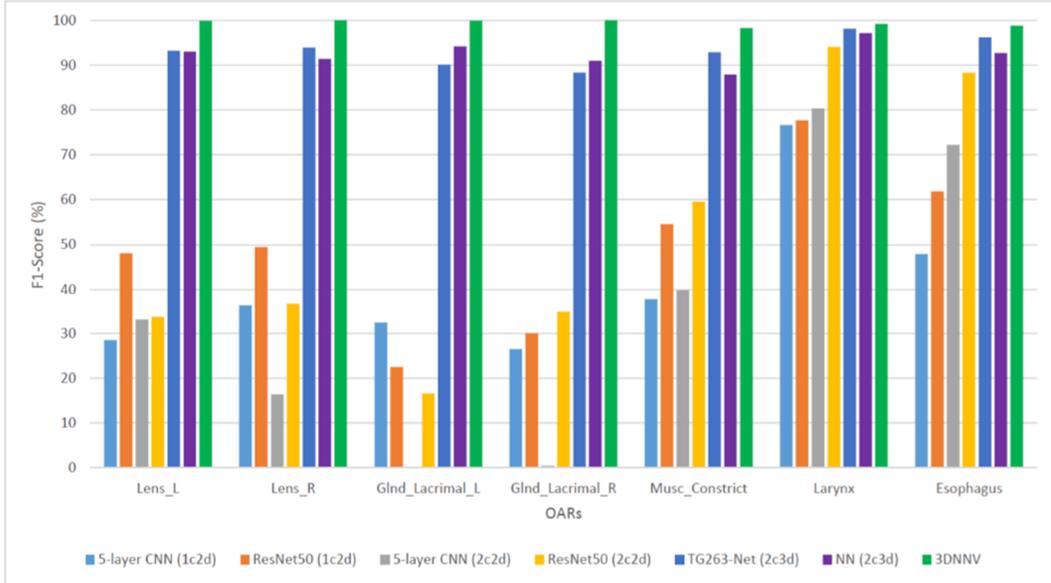

**FIGURE 6.** Average F1 Scores (%) for deep learning-based models' classification of small-volume OARs in HN_UTSW.

Based on the results of the above experiments, we set three more architectures—TG263-Net (2c3d), NN (2c3d), and 3DNNV—to determine the optimal sampling strategy and structure for the framework. For TG263-Net [20], we loosely used the encoder in V-Net [44] to construct the classifier. Then, we normalized the voxel size of CT and mask volumes in raw data to 2 mm : 2 mm : 2 mm, and we cropped the central 64 × 64 × 64 cubes (on CT and mask) to construct the 2-channel input. We randomly translated the center-of-mass by 10 mm to gain 9 inputs for each sample. In the inference phase, the 9 vectors extracted from 9 inputs vote for a final prediction result. This sampling strategy is similar to 3DNNV's, so we applied the TG263-Net's sampling strategy (along with the voting strategy) to our Non-local Network (NN) and designated the architecture as NN (2c3d). When we compared these two architectures, NN (2c3d) performed notably worse than TG263-Net (2c3d). Nevertheless, after replacing the NN (2c3d)'s sampling and voting strategy with ASAC/Voting, we arrived at the framework of 3DNNV, which includes the improved data processing strategy and the optimized feature extraction structure. The average TPR, F1, and AUC for 3DNNV and the other DL-based methods over all categories on all test datasets are shown in Table VI. Although TG263-Net performed slightly better than NN, it required a longer running time. Most importantly, 3DNNV outperformed the other DL-based methods and had better generalizability across institutional datasets.

TABLE V
AVERAGE F1 SCORE (%) OF 3DNNV AND OTHER DL-BASED METHODS FOR PITUITARY AND OPTICCHIASM ON HN_UTSW.

| Architecture | Params (M) | FLOPs (G) | Pituitary | OpticChiasm |
|---|---|---|---|---|
| 5-layer CNN [17] (1c2d) | 34.71 | 1.74 | 0.00 ± 0.00 | 30.61 ± 5.49 |
| ResNet50 [25] (1c2d) | 24.03 | 1.32 | 0.00 ± 0.00 | 39.55 ± 5.30 |
| 5-layer CNN [17] (2c2d) | 34.71 | 1.74 | 0.00 ± 0.00 | 3.49 ± 4.81 |
| ResNet50 [25] (2c2d) | 24.03 | 1.32 | 7.99 ± 5.84 | 58.44 ± 16.88 |
| TG263-Net [20] (2c3d) | 48.91 | 25.36 | 52.07 ± 18.41 | 92.92 ± 2.75 |
| NN (2c3d) | 69.83 | 7.85 | 20.78 ± 10.03 | 75.96 ± 12.59 |
| 3DNNV | 69.83 | 7.24 | **91.17±3.18** | **99.47±0.18** |

TABLE VI
PERFORMANCE RESULTS OF 3DNNV AND OTHER DL-BASED METHODS ACROSS THREE DATASETS.

| Architecture | HN_PETCT | | | PDDCA | | | HN_UTSW | | |
|---|---|---|---|---|---|---|---|---|---|
| | TPR | F1 | AUC | TPR | F1 | AUC | TPR | F1 | AUC |
| 5-layer CNN [17] (1c2d) | 74.28±22.88 | 73.57±23.20 | 86.82±11.46 | 58.61±36.58 | 62.65±33.63 | 78.78±18.27 | 57.10±25.63 | 55.62±22.34 | 77.87±12.61 |
| ResNet50 [25] (1c2d) | 74.86±25.18 | 75.08±25.21 | 87.13±12.61 | 60.37±36.55 | 62.92±32.06 | 79.45±18.02 | 60.41±28.54 | 56.42±21.88 | 79.55±13.98 |
| 5-layer CNN [17] (2c2d) | 79.76±28.13 | 79.48±27.87 | 89.68±14.16 | 93.04±9.65 | 95.60±5.46 | 96.45±4.81 | 44.89±37.10 | 39.26±31.45 | 71.53±17.56 |
| ResNet50 [25] (2c2d) | 86.97±22.04 | 86.48±21.62 | 93.37±11.01 | 94.32±9.91 | 96.09±6.53 | 97.10±4.93 | 74.98±26.07 | 73.33±26.48 | 87.19±12.93 |
| TG263-Net [20] (2c3d) | 97.84±5.34 | 97.08±6.96 | 98.90±2.69 | 95.76±11.49 | 97.15±7.91 | 97.85±5.82 | 95.99±6.52 | 94.46±9.01 | 97.95±3.26 |
| NN (2c3d) | 94.27±14.17 | 91.72±15.85 | 97.02±7.09 | 89.12±31.10 | 89.78±29.86 | 94.55±15.58 | 91.36±12.83 | 89.58±18.19 | 95.58±6.49 |
| 3DNNV | **99.81±0.63** | **99.82±0.39** | **99.90±0.31** | **100±0** | **100±0** | **100±0** | **98.87±2.37** | **98.90±1.91** | **99.42±1.19** |

### D. 3DNNV's extensibility

To demonstrate the 3DNNV's extensibility, we fine-tuned the model on other anatomical sites.

Data from 8 lung region patients and 5 prostate region patients were selected to fine-tune the model: we used the parameters of 3DNNV pre-trained on the 28 head-and-neck OAR data for initialization, then we froze all parameters except those on the fourth residual block (*Res4* shown in Fig. 3) and the fully-connected layer, and we set the learning rate as 1e-5 for the trainable layers. Next, we tested the fine-tuned model on data from 29 lung region patients and 28 prostate region patients. Other training settings were the same as for 3DNNV.

TABLE VII
TESTING THE FINE-TUNED MODEL ON OTHER ANATOMICAL SITES (LUNGS AND PROSTATE).

| Region | OARs | Train (samples) | Test (samples) | TPR (epoch=5) | TPR (epoch=7) | TPR (epoch=10) | TPR (epoch=20) |
|---|---|---|---|---|---|---|---|
| Lungs | Heart | 8 | 30 | 63.33% | 100.00% | 100.00% | 100.00% |
| | Carina | 6 | 31 | 100.00% | 100.00% | 100.00% | 100.00% |
| | Lungs | 6 | 23 | 86.96% | 100.00% | 100.00% | 100.00% |
| | Lung_L | 6 | 26 | 3.85% | 84.62% | 96.15% | 96.15% |
| | Lung_R | 6 | 27 | 11.11% | 96.30% | 96.30% | 96.30% |
| | Skin | 2 | 3 | 100.00% | 100.00% | 100.00% | 100.00% |
| Prostate | Rectum | 5 | 30 | 100.00% | 100.00% | 100.00% | 100.00% |
| | Bladder | 5 | 29 | 89.66% | 96.55% | 96.55% | 96.55% |
| | Sigmoid | 4 | 21 | 19.05% | 57.14% | 71.43% | 80.95% |
| | PenileBulb | 4 | 25 | 52.00% | 100.00% | 100.00% | 100.00% |
| | Femurs | 3 | 12 | 100.00% | 100.00% | 100.00% | 100.00% |
| | Femur_Head_L | 2 | 20 | 100.00% | 100.00% | 100.00% | 100.00% |
| | Femur_Head_R | 2 | 20 | 100.00% | 100.00% | 100.00% | 100.00% |

The experimental results are shown in Table VII. The model only needed 20 epochs to transfer to recognizing OARs in other anatomical regions with a small amount of data, and it obtained a good recognition accuracy. This means that, with very little data and a short amount of time, we can easily transfer the pre-trained model to the target anatomical sites to meet the needs of the new application.

## V. DISCUSSION

### A. Effectiveness

As mentioned before, 3DNNV yields better performance at identifying small-volume and error-prone OARs than all other deep learning-based models we investigated. To some degree, the sampling/voting strategies applied in TG263-Net and our framework are similar: generate many inputs for a sample, and vote for a final result at the inference phase. Here, we try to explain why 3DNNV works for error-prone OARs, and compare it with TG263-Net. The 256-d vectors (outputs of the network) for error-prone OARs are visualized in Fig. 7. There are partial small-volume OARs in the head-and-neck region, the data of which are often poorly delineated and imbalanced; some of these small-volume OARs are similar in location and shape, such as Pituitary and OpticChiasm. Fig. 7 (a) and Fig. 7 (b) indicate the predictive results of TG263-Net on small-volume and error-prone OARs; apparently, most of these OARs are hard to identify without the voting strategy. However, after applying the voting strategy, the OARs that come with similar shapes/locations/sizes and imbalanced training samples still tend to be confused, like Pituitary and OpticChiasm. 3DNNV's improved data processing strategy and optimized feature extract structure solved this problem, as shown in Fig. 7 (c) and Fig. 7 (d). We gained more reliable and credible results: clear boundaries between different categories allow easier classification.

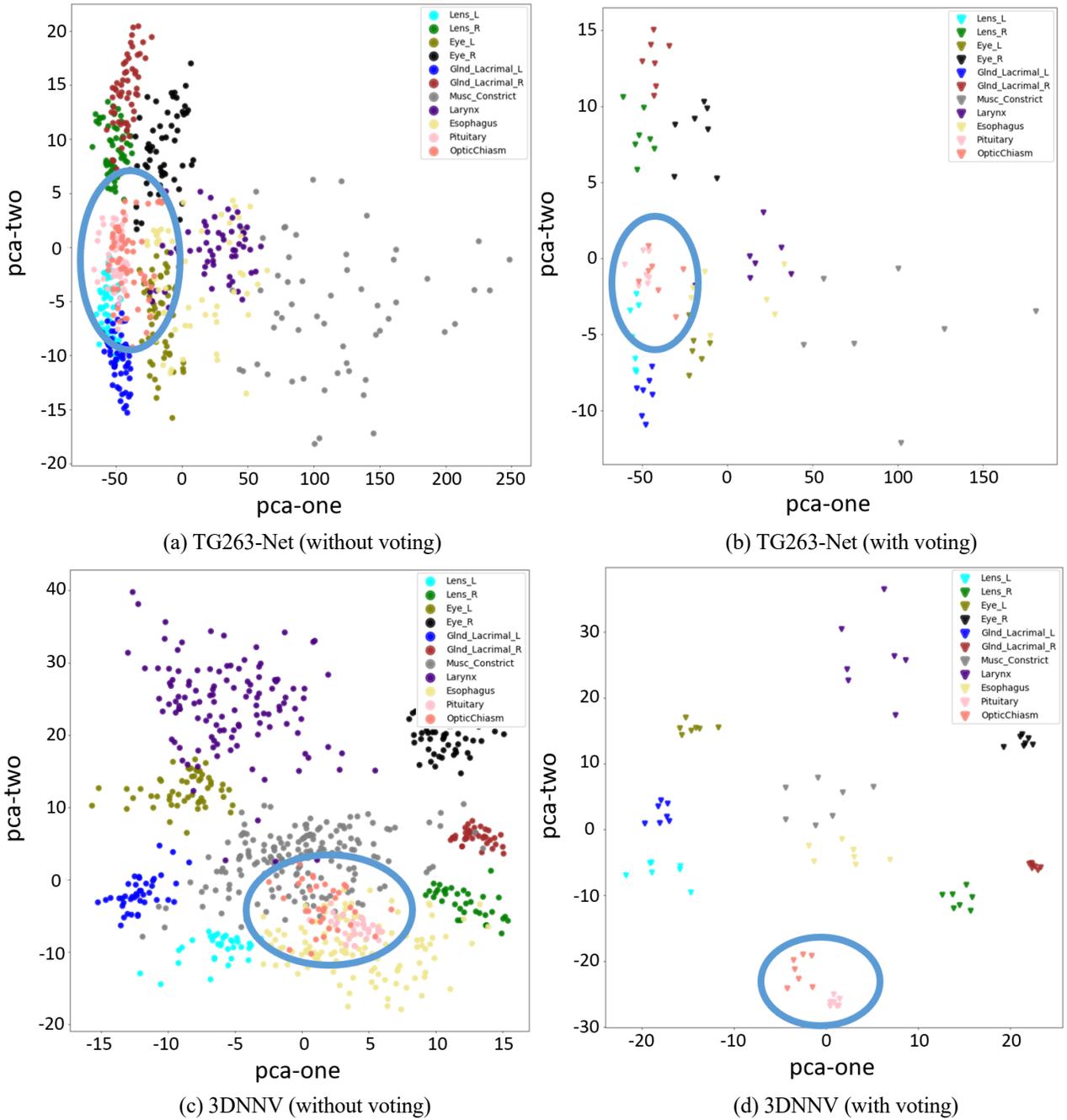

**FIGURE 7.** Visualization of the predictive results on the test dataset (HN_UTSW). To show the performance of 3DNNV, we compared it with TG263-Net [20]. For each category of small-volume OARs shown in the top-right legend, 9 samples were selected from dataset HN_UTSW and fed into networks to extract high-level features (256-d vectors). Then, we reduced the dimensionality of the high-level features by using Principal Component Analysis (PCA) [53], and the result is illustrated in the figure. We highlighted the results of classifying Pituitary and OpticChiasm. (a) and (b) are the results of TG263-Net, which still confused Pituitary with OpticChiasm. (c) and (d) show the clear boundaries between different small-volume OARs when using the ASAC/Voting strategy.

### B. Statistical significance of the performance improvement

To illustrate the statistical significance of 3DNNV's performance improvement over the Baseline (GS) model, we performed a one-way analysis of variance (ANOVA) test on the results of the Baseline (GS) and 3DNNV models over all 28 categories of OARs in the head-and-neck datasets. The mean difference denotes the difference between the average TPR/F1/AUC values over six sets of models tested on the datasets (Table VIII). Positive numbers in the mean difference indicate that 3DNNV performed better than Baseline (GS), and negative numbers indicate that 3DNNV performed worse. We set the $p$-value as 1.0 for samples whose variance were identical between the Baseline (GS) and 3DNNV. We found that 3DNNV significantly outperformed Baseline (GS) ($p$-value $< 0.05$, $Mean > 0.0$) in identifying small-volume and error-prone OARs, especially in HN_UTSW (Table VIII).

TABLE VIII
ONE-WAY ANALYSIS OF VARIANCE (ANOVA). TO ILLUSTRATE THE IMPROVEMENT DIRECTLY, WE COMPARE THE RESULTS OF 3DNNV WITH BASELINE (GS) IN TERMS OF TPR, F1, AND AUC. BOLDFACE INDICATES STATISTICALLY SIGNIFICANT IMPROVEMENT (THRESHOLD P-VALUE $< 0.05$, THE MEAN DIFFERENCE $> 0$).

| OARs | HN_PETCT | | | | | | PDDCA | | | | | | HN_UTSW | | | | | |
|---|---|---|---|---|---|---|---|---|---|---|---|---|---|---|---|---|---|---|
| | TPR | | F1 | | AUC | | TPR | | F1 | | AUC | | TPR | | F1 | | AUC | |
| | Mean | p-value | Mean | p-value | Mean | p-value | Mean | p-value | Mean | p-value | Mean | p-value | Mean | p-value | Mean | p-value | Mean | p-value |
| Lens_L | 10.34% | **0.0108** | 8.09% | **< 0.01** | 5.26% | **< 0.01** | | | | | | | 16.45% | **0.0171** | 10.67% | **0.0158** | 8.24% | **0.0169** |
| Lens_R | 1.15% | 0.1449 | 2.54% | **< 0.01** | 0.64% | 0.1054 | | | | | | | 6.28% | **< 0.01** | 6.25% | **< 0.01** | 3.19% | **< 0.01** |
| Eye_L | 0.00% | 1 | -0.92% | 1 | -0.06% | 1 | | | | | | | 0.46% | 0.0856 | 5.69% | **< 0.01** | 0.40% | **< 0.01** |
| Eye_R | -1.85% | 1 | -0.48% | **0.0425** | -0.90% | **< 0.01** | | | | | | | 0.83% | **0.0346** | 4.42% | **< 0.01** | 0.53% | **< 0.01** |
| Glnd_Lacrimal_L | 50.00% | **< 0.01** | 52.42% | **< 0.01** | 25.12% | **< 0.01** | | | | | | | 24.77% | **< 0.01** | 18.62% | **< 0.01** | 12.50% | **< 0.01** |
| Glnd_Lacrimal_R | 55.56% | **< 0.01** | 43.33% | **< 0.01** | 27.80% | **< 0.01** | | | | | | | 16.52% | **<0.01** | 10.72% | **< 0.01** | 8.30% | **< 0.01** |
| Parotid_L | 0.00% | 1 | 0.14% | 0.3409 | 0.01% | 0.3409 | 0.00% | 1 | 0.00% | 1 | 0.00% | 1 | 0.46% | **0.0107** | 0.26% | 0.1961 | 0.23% | **0.0111** |
| Parotid_R | 0.00% | 1 | 0.00% | 1 | 0.00% | 1 | 0.35% | 0.3409 | 0.18% | 0.3409 | 0.17% | 0.3409 | 0.19% | 0.3409 | 0.47% | 0.3409 | 0.12% | 0.2238 |
| Glnd_Submand_L | 0.76% | 0.3409 | 4.35% | **< 0.01** | 0.48% | 0.2304 | 0.00% | 1 | 0.00% | 1 | 0.00% | 1 | 3.30% | **< 0.01** | 3.37% | **< 0.01** | 1.72% | **< 0.01** |
| Glnd_Submand_R | 0.00% | 1 | 0.00% | 1 | 0.00% | 1 | 0.00% | 1 | 0.23% | 0.3409 | 0.02% | 0.3409 | 0.99% | **< 0.01** | 1.63% | **< 0.01** | 0.54% | **< 0.01** |
| Cavity_Oral | 2.60% | 0.1114 | 3.61% | **0.0206** | 1.39% | 0.0976 | | | | | | | 8.80% | **< 0.01** | 5.13% | **< 0.01** | 4.43% | **< 0.01** |
| Lips | 15.08% | **< 0.01** | 9.63% | **< 0.01** | 7.57% | **< 0.01** | | | | | | | 0.00% | 1 | -2.30% | **< 0.01** | -0.01% | **< 0.01** |
| Bone_Mandible | 0.00% | 1 | 1.83% | **0.0268** | 0.12% | **0.0273** | 0.00% | 1 | 0.00% | 1 | 0.00% | 1 | 1.12% | **< 0.01** | 0.83% | **< 0.01** | 0.58% | **< 0.01** |
| Cochlea_L | 0.00% | 1 | 0.00% | 1 | 0.00% | 1 | | | | | | | 2.72% | **< 0.01** | 1.49% | **< 0.01** | 1.37% | **< 0.01** |
| Cochlea_R | 0.00% | 1 | 0.00% | 1 | 0.00% | 1 | | | | | | | 1.44% | **0.0372** | 1.04% | **< 0.01** | 0.74% | **0.0334** |
| Musc_Constrict | -1.92% | 0.0924 | -0.67% | 0.2770 | -0.95% | 0.0947 | | | | | | | 19.76% | **< 0.01** | 15.93% | **< 0.01** | 10.02% | **< 0.01** |
| Larynx | 14.18% | **< 0.01** | 9.28% | **< 0.01** | 7.17% | **< 0.01** | | | | | | | 4.64% | **< 0.01** | 3.16% | **< 0.01** | 2.36% | **< 0.01** |
| Esophagus | 2.50% | **0.0493** | 1.28% | 0.0593 | 1.25% | **0.048** | | | | | | | 4.70% | **< 0.01** | 2.67% | **< 0.01** | 2.36% | **< 0.01** |
| BrachialPlex_L | 2.78% | 0.1739 | 2.36% | 0.1145 | 1.41% | 0.1710 | | | | | | | 1.20% | **< 0.01** | 0.73% | **< 0.01** | 0.61% | **< 0.01** |
| BrachialPlex_R | 0.00% | 1 | 0.45% | 0.3409 | 0.01% | 0.3409 | | | | | | | 0.57% | **< 0.01** | 0.42% | **< 0.01** | 0.29% | **< 0.01** |
| Thyroid | 6.06% | 0.0734 | 19.19% | **< 0.01** | 3.30% | 0.0637 | | | | | | | -5.56% | 0.0924 | 3.89% | **0.0267** | -2.77% | 0.0936 |
| Brain | 0.00% | 1 | 0.00% | 1 | 0.00% | 1 | | | | | | | -1.60% | **< 0.01** | 16.95% | **< 0.01** | -0.58% | **< 0.01** |
| Brainstem | 0.00% | 1 | 0.00% | 1 | 0.00% | 1 | 0.00% | 1 | 0.00% | 1 | 0.00% | 1 | 1.51% | 0.1620 | 1.91% | **0.0191** | 0.84% | 0.1303 |
| Pituitary | 62.50% | **< 0.01** | 60.66% | **< 0.01** | 31.38% | **< 0.01** | | | | | | | 33.33% | **< 0.01** | 60.72% | **< 0.01** | 16.84% | **< 0.01** |
| OpticChiasm | 11.11% | **< 0.01** | 11.23% | **< 0.01** | 5.71% | **< 0.01** | 16.32% | **0.0177** | 10.50% | **0.0206** | 8.30% | **0.0163** | 8.26% | **< 0.01** | 13.00% | **< 0.01** | 4.35% | **< 0.01** |
| OpticNrv_L | 0.43% | 0.3409 | 1.05% | 0.0647 | 0.25% | 0.2737 | 2.08% | 1 | 1.05% | 1 | 1.04% | 1 | 0.25% | 0.4480 | 4.25% | **< 0.01** | 0.24% | 0.1878 |
| OpticNrv_R | 0.43% | 0.3409 | 0.22% | 0.3409 | 0.21% | 0.3409 | 2.78% | **< 0.01** | 1.41% | **< 0.01** | 1.39% | **< 0.01** | 1.14% | **< 0.01** | 2.96% | **0.016** | 0.63% | **< 0.01** |
| SpinalCord | 0.34% | 0.3409 | 0.84% | **0.0212** | 0.0021 | 0.2271 | | | | | | | 1.25% | **< 0.01** | 2.98% | **< 0.01** | 0.78% | **< 0.01** |

*C. Limitations*

1) RUNNING TIME

To reduce the running time and improve the performance of 3DNNV, we added an early-match module (Fig. 8) to the framework and maintained a locally standardized label dictionary. The early-match module performs string matching [54] between the original label and the standardized label: if and only if the original label fully matches one of the standardized labels in the dictionary, then the standardized label is used to rename the given structure. This reduces the number of structures to be processed by 3DNNV and allows the framework to process unknown structures not included in the training dataset.

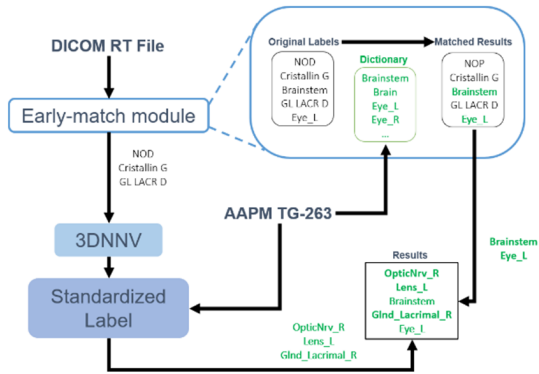

**FIGURE 8.** Diagram of the early-match module.

Originally, 3DNNV was used to process a patient data containing 38 structures. A running time of 7 $m$ 41.83 $s$ was required to obtain all the recognition results. This running time is too long to be acceptable for further applications. To solve this problem, we added the early-match module before feeding the input into the 3DNNV. This module relies on a pre-stored dictionary as the basis for string matching. After adding the early-match module to the framework, 3DNNV only needed to process 17 OARs in this patient (containing 38 structures), so the total running time was 3 $m$ 36.05 $s$. Timely updates and maintenance for the dictionary will help to optimize the automatic identification process and avoid reprocessing labels that have already been standardized. However, the limitation is that the dictionary can only handle one-to-one mapping. When given RT data collected from a multi-language environment, the dictionary mapping method will not significantly reduce the running time of standardization. This is why a single-dictionary mapping method cannot handle cross-institutional data.

2) MULTIPLE LABELS FOR THE SAME STRUCTURE

The original 3DNNV model was trained and tested on only 28 OARs in head-and-neck datasets, which limits the model's recognition range to these 28 categories. To make the model generalizable to more structures, we tried to extend it to other anatomical sites, and it worked well. However, like Schuler et al. [8], we found that the model cannot distinguish typographic name variations from fundamental semantic differences in the same structure. In this work, we mainly discuss standardizing OAR labels, but in practice, the structures in individual RT data will be labeled differently for different treatment purposes. For example, the same structures might be labeled CavityOral_avoid or CavityOral, SpinalCord or SpinalCord_5mm, IL_Parotid, CL_Parotid, Parotid_L, or Parotid_R, depending on the specific application for which the labels are being used. These inputs have similar semantic features in images, so it is very difficult to identify these structures based on image information. At the same time, some non-target structures will have multi-level labels for a single OAR—such as Musc_Constrict_M, Musc_Constrict_S, Musc_Constrict_I, and Musc_Constrict, or OpticChiasm_aaa and OpticChiasm_bbb, where aaa is the resident's name and bbb is the actual attending physician's name—depending on different RT plans and local policies. These standardization conventions may vary across different medical institutions and treatment plans. At the same time, the standardization of target volumes also warrants attention. The target volume often overlaps with OARs and could be misidentified as an OAR. Additional information can be used to help identifying target volumes, such as positron emission tomography (PET), which is widely used in the clinical practice and able to accurately define biological target volume (BTV). The utilization of BTV and gross tumor volume (GTV) will improve the accuracy for identifying clinical target volume (CTV) [55, 56]. Adding text information may also help us to improve the performance of 3DNNV and meet the requirements of clinical applications.

3) OUTLIERS

In previous experiments, we found that the masks collected from different clinical centers may have inconsistent contours. These inconsistencies result from differences in physician experience and in how the local institution defines delineation for OARs. Moreover, there are outliers in many datasets: some lack masks in some slices; in other cases, the label does not always match the contour in the mask because of inaccurate delineation or partial depiction. We believe that detecting delineation outliers also presents a challenge to standardizing nomenclature for RT data.

## VI. CONCLUSIONS

In this paper, we propose a novel framework, 3DNNV, that combines an ASAC/Voting strategy and a non-local network to integrate clinicians' domain knowledge and recognition mechanisms into our deep learning architecture. To the best of our knowledge, our work is the first to propose an architecture that integrates domain knowledge to solve the recognition problems caused by imbalance and poor delineation. Our model had a significantly higher average true positive rate than the baseline model across three test datasets (+ 8.27%, + 2.39%, and + 5.53%). More importantly, our model outperformed the baseline in terms of the F1 Score of the Pituitary (28.63% to 91.17%) with only 9 training samples, when tested on the HN_UTSW dataset.

We visualized the vectors of our predictive results to evaluate the effectiveness of 3DNNV. One-way ANOVA tests showed the statistical significance of 3DNNV's performance improvement over Baseline (GS). Finally, we discussed limitations of the model that could impede application, and we suggested future work for automatically standardizing anatomical structure nomenclature in radiotherapy.

Our findings in this work will advance efforts to automate the standardization of organ labels in DICOM RT data, which will facilitate and improve data-driven research.

## ACKNOWLEDGEMENT

We would like to thank Dr. Jonathan Feinberg for editing the manuscript.

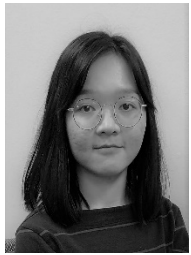

**Qiming Yang** received a B.S. degree in software engineering from the Sun Yat-sen University, Guangzhou, China, in 2017. Her previous works focused on biomedical image processing. She is currently pursuing an M.S. degree in Sun Yat-sen University and was visiting the University of Texas Southwestern Medical Center in Dallas, Texas, from September 2018 to August 2019. Her research interests include image processing, deep learning, and computer vision.

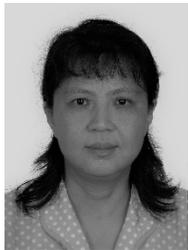

**Hongyang Chao** received B.S. and Ph.D. degrees in computational mathematics from Sun Yat-sen University, Guangzhou, China. In 1988, she joined the Department of Computer Science, Sun Yat-sen University, where she was initially an Assistant Professor and later became an Associate Professor. She is currently a Full Professor in the School of Data and Computer Science. She has published extensively in the area of image/video processing and holds 3 U.S. patents and 4 Chinese patents in the related area. Her current research interests include image and video processing, image and video compression, massive multimedia data analysis, and content-based image (video) retrieval. She was visiting the University of Texas Southwestern Medical Center in Dallas, Texas, from September 2018 to August 2019.

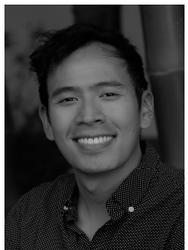

**Dan Nguyen** is currently an Assistant Professor in the Medical Artificial Intelligence and Automation (MAIA) Laboratory at the University of Texas Southwestern Medical Center in Dallas, Texas. He received a B.S. in Physics at the University of Texas at Austin in 2012 and a Ph.D. in Biomedical Physics at the University of California, Los Angeles in 2017. His current research in MAIA Lab includes using artificial intelligence technologies and advanced optimization algorithms for radiation therapy treatment planning. In particular, he is tackling problems involving clinical volumetric dose prediction, Pareto surface navigation, incorporating human and learned domain knowledge, dose calculation, beam orientation optimization, and uncertainty estimation.

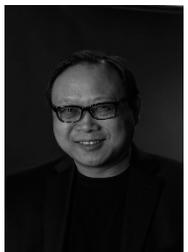

**Steve Jiang** received his Ph.D. in Medical Physics from Medical College of Ohio in 1998. After completing his postdoctoral training at Stanford University, he joined Massachusetts General Hospital and Harvard Medical School in 2000 as an Assistant Professor of Radiation Oncology. In 2007, Dr. Jiang was recruited to University of California San Diego as a tenured Associate Professor to build the Center for Advanced Radiotherapy Technologies, for which he was the founding and executive director. He was then promoted to Full Professor with tenure in 2011. In October 2013, Dr. Jiang joined University of Texas Southwestern Medical Center as a tenured Full Professor, Barbara Crittenden Professor in Cancer Research, Vice Chair of Radiation Oncology Department, and Director of Medical Physics and Engineering Division. He is the founding director of the Medical Artificial Intelligence and Automation Laboratory. Dr. Jiang is a Fellow of Institute of Physics and American Association of Physicists in Medicine. His current research interest is to develop and deploy artificial intelligence technologies to solve medical problems.